\documentclass[letterpaper, 10 pt, conference]{ieeeconf}  

\IEEEoverridecommandlockouts                              

\usepackage{graphics} 
\usepackage{booktabs}
\usepackage{subfigure}
\usepackage{graphicx}
\usepackage{multirow}
\usepackage[table]{xcolor}
\usepackage{colortbl}
\usepackage{amsmath,amssymb,amsfonts}
\usepackage{cite}
\usepackage{tabulary}
\usepackage{tabularray}
\usepackage{tikz}
\usepackage{color}
\usepackage[colorlinks=false, pdfborder={0 0 0}]{hyperref}

\newcommand\crule[3][black]{\textcolor{#1}{\rule{#2}{#3}}}
\definecolor{roadcolor}{RGB}{234,51,246}
\definecolor{sidewalkcolor}{RGB}{68,8,72}
\definecolor{parkingcolor}{RGB}{241,156,249}
\definecolor{othergroundcolor}{RGB}{160,32,76}
\definecolor{buildingcolor}{RGB}{246,202,69}
\definecolor{carcolor}{RGB}{111,149,238}
\definecolor{truckcolor}{RGB}{74,32,172}
\definecolor{bicyclecolor}{RGB}{136,227,242}
\definecolor{motorcyclecolor}{RGB}{37,59,146}
\definecolor{othervehiclecolor}{RGB}{96,81,242}
\definecolor{vegetationcolor}{RGB}{79, 173, 50}
\definecolor{trunkcolor}{RGB}{126, 65, 22}
\definecolor{terraincolor}{RGB}{171, 238, 105}
\definecolor{personcolor}{RGB}{234, 60, 49}
\definecolor{bicyclistcolor}{RGB}{234, 66, 195}
\definecolor{motorcyclistcolor}{RGB}{138, 42, 90}
\definecolor{fencecolor}{RGB}{238, 128, 69}
\definecolor{polecolor}{RGB}{252, 241, 161}
\definecolor{trafficsigncolor}{RGB}{233, 51, 35}
\definecolor{color1}{RGB}{176, 36, 24}
\definecolor{color2}{RGB}{119,185,0}
\definecolor{color3}{RGB}{0, 0, 200}
\definecolor{colorofteaser}{RGB}{176, 36, 24}
\definecolor{color4}{RGB}{0, 0, 0}

\title{
\LARGE \bf
A Coarse-to-Fine Approach to Multi-Modality \\3D Occupancy Grounding
}

\author{Zhan Shi$^{1}$, Song Wang$^{1}$, Junbo Chen$^{2}$ and Jianke Zhu$^{1}$, \textit{Senior Member, IEEE}
\thanks{$^{1}$Zhan Shi is with the College of Software Technology, Zhejiang University. Song Wang and Jianke Zhu are with the College of Computer Science, Zhejiang University, Hangzhou 310027, China. Jianke Zhu is the corresponding author. (email: \{zscoisini,songw, jkzhu\}@zju.edu.cn).}%
\thanks{$^{2}$Junbo Chen is with the Udeer.ai, Hangzhou 310000, China. (email: junbo@udeer.ai).}
}

\makeatletter
\renewcommand{\paragraph}{%
  \@startsection{paragraph}{4}%
  {\z@}{1.1ex \@plus 1ex \@minus .2ex}{-1mm}%
  {\normalfont\normalsize\bfseries}%
}
\makeatother

\begin{document}
\maketitle
\thispagestyle{empty}
\pagestyle{empty}

\begin{abstract}
Visual grounding aims at identifying objects or regions in a scene based on natural language descriptions, which is essential for spatially aware perception in autonomous driving. However, existing visual grounding tasks typically depend on bounding boxes that often fail to capture fine-grained details. 
Not all voxels within a bounding box are occupied, resulting in inaccurate object representations.
To address this, we introduce a benchmark for 3D occupancy grounding in challenging outdoor scenes. 
Built on the nuScenes dataset, it fuses natural language with voxel-level occupancy annotations, offering more precise object perception compared to the traditional grounding task.
Moreover, we propose GroundingOcc, an end-to-end model designed for 3D occupancy grounding through multimodal learning. It combines visual, textual, and point cloud features to predict object location and occupancy information from coarse to fine. 
Specifically, GroundingOcc comprises a multimodal encoder for feature extraction, an occupancy head for voxel-wise predictions, and a grounding head for refining localization. 
Additionally, a 2D grounding module and a depth estimation module enhance geometric understanding, thereby boosting model performance.
Extensive experiments on the benchmark demonstrate that our method outperforms existing baselines on 3D occupancy grounding.
The dataset is available at \url{https://github.com/RONINGOD/GroundingOcc}.
\end{abstract}    
\section{Introduction}
\label{sec:intro}
\begin{figure}[!t]
    \centering
    \setlength{\abovecaptionskip}{0pt}
    \vspace{1.5mm}
    \includegraphics[width=\columnwidth]{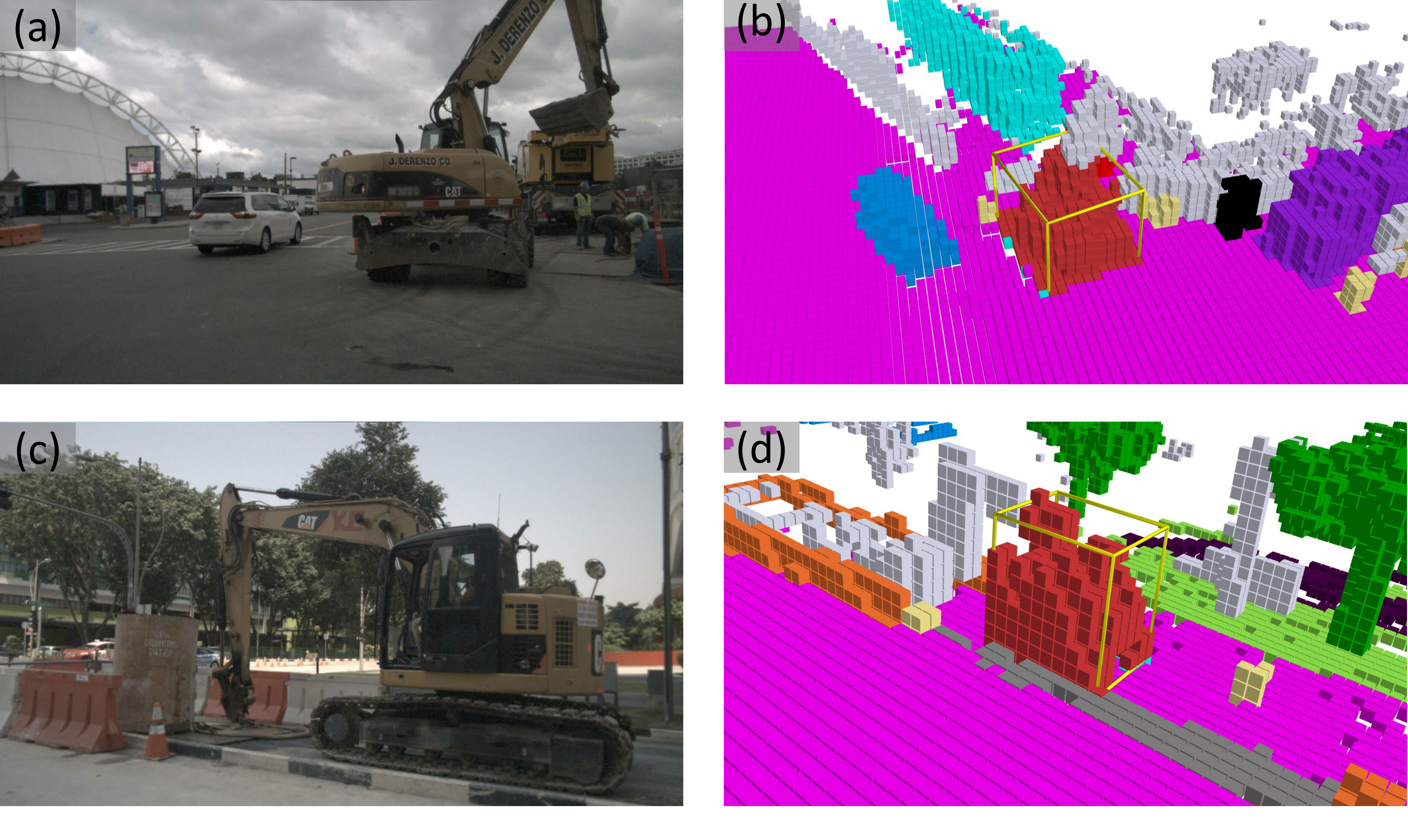} 
    \caption{\textbf{Examples of 3D occupancy grounding.} Conventional bounding boxes cannot accurately describe irregularly shaped vehicles in daily driving scenes, such as the excavators in (a) and (c). In contrast, 3D occupancy representations provide a more precise depiction of complex obstacles, as shown in (b) and (d).}
    \label{fg:example of occ grounding}
    \vspace{-6mm}
\end{figure}

Holistic 3D scene understanding is a key challenge in computer vision, especially for autonomous vehicles \cite{arnold2019survey, hu2023planning}. 
However, current research~\cite{vobecky2024pop,tan2023ovo,yu2025inst3d} struggles with fine-grained spatial reasoning and cognitive scene interpretation.
Visual grounding~\cite{zhang2018grounding} is one of the most promising tasks in this domain, which focuses on localizing objects from natural language descriptions. 
Traditional bounding box-based methods~\cite{zhan2024mono3dvg,luo20223d} often fail to capture full complexity of real-world objects. This is particularly evident for irregularly shaped or partially occluded objects~\cite{tong2023scene}. 
As illustrated in Fig. \ref{fg:example of occ grounding}, bounding boxes are insufficient for accurately representing complex objects, such as excavators, with non-rectangular structures. 
In contrast, 3D occupancy representations provide a more precise way to capture the object shapes and details.

\begin{figure*}[!htbp]
  \centering
  \setlength{\abovecaptionskip}{0pt}
  \includegraphics[width=0.95\textwidth]{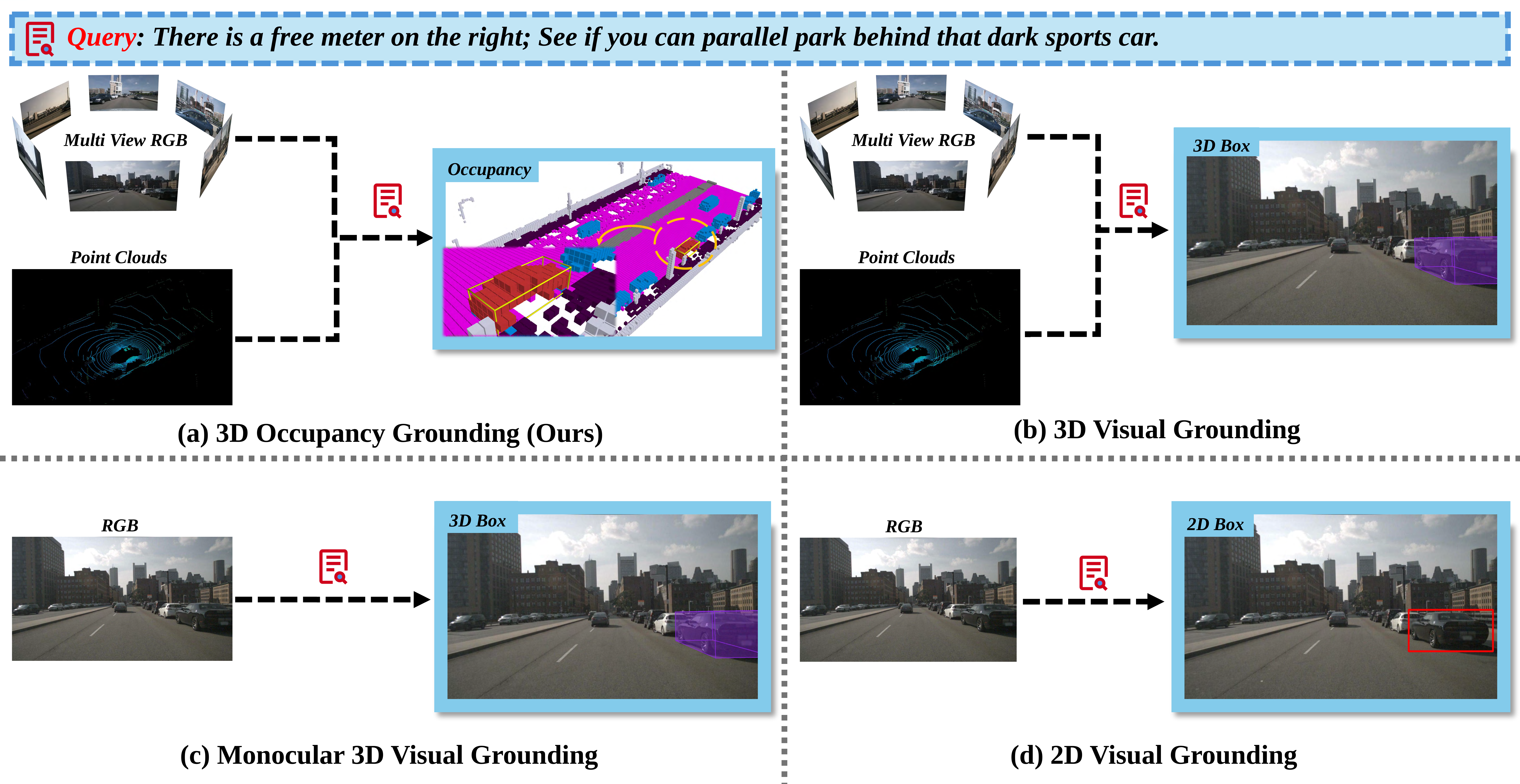}
  \caption{\textbf{Illustration for 3D occupancy grounding.} 
  (a) Our proposed 3D occupancy grounding aims to perceive the 3D occupancy of the referred object in a scene using language descriptions. 
  (b) The traditional 3D visual grounding intends to estimate the location and size of the referred object. 
  (c) Monocular 3D visual grounding relies only on a single image to localize the 3D extent of the referred object.
  (d) The counterpart 2D task does not capture the 3D extent of the referred object.
  }
  \label{fig:task-introduce}
  \vspace{-5mm}
\end{figure*}

Existing datasets have made significant progress in visual grounding, with ScanRefer \cite{chen2020scanrefer} and Sr3D \cite{achlioptas2020referit3d} used for indoor object localization, and datasets such as Mono3DRefer \cite{zhan2024mono3dvg} and Talk2Car \cite{deruyttere2019talk2car} for autonomous driving. Emerging datasets such as Talk2Radar \cite{guan2024talk2radar} and Talk2LiDAR \cite{talk2lidar} contribute but are not publicly available. Moreover, these datasets are limited to bounding box-based understanding and do not incorporate voxel-level occupancy prediction, which is essential for fine-grained spatial understanding.

We bridge this gap by integrating occupancy prediction into 3D visual grounding. As shown in Fig. \ref{fig:task-introduce}, our 3D occupancy grounding task combines natural language descriptions with voxel-level spatial precision, allowing a more detailed understanding of complex environments. This task enhances both localization and fine-grained occupancy perception, which are crucial for autonomous decision-making.

To advance this research, we introduce Talk2Occ, a new benchmark designed for 3D occupancy grounding in autonomous driving scenarios. Talk2Occ combines the Talk2Car \cite{deruyttere2019talk2car} dataset with the Occ3D \cite{tian2024occ3d} dataset, linking them through the nuScenes dataset~\cite{caesar2020nuscenes}. 
The Talk2Occ dataset includes surround-view images, LiDAR point clouds, 3D occupancy ground truth, and corresponding natural language descriptions. This new task presents unique challenges, including the need for more precise localization and detailed spatial understanding, which motivates the development of advanced models for 3D occupancy grounding.

To address the challenges of 3D occupancy grounding, we propose GroundingOcc, an end-to-end model that integrates occupancy prediction with grounding. Our approach combines 2D grounding and a depth estimation module, which is supervised by occupancy-rendered depth maps, enabling it to utilize spatial information from multiple modalities.

Our main contributions are summarized as below.
\begin{itemize} 
\item We introduce occupancy prediction into 3D visual grounding to enable text-guided human-machine interaction for fine-grained occupancy perception.
\item A novel benchmark for 3D occupancy grounding in autonomous driving, Talk2Occ, bridges natural language with voxel-level spatial understanding.
\item An end-to-end multi-branch network, GroundingOcc, that integrates visual, textual, and point-cloud features for precise voxel-level occupancy grounding. It leverages a multi-branch architecture with 2D grounding and depth estimation to improve accuracy.
\item Our framework achieves state-of-the-art performance in 3D occupancy grounding and sets a new benchmark for voxel-level perception. Extensive experiments demonstrate that our method significantly improves localization accuracy, outperforming the baseline by 18.13\%.
\end{itemize}
\section{Related Work}
\label{sec:related}
\noindent \textbf{2D Visual Grounding.}  
Localizing the referred objects in 2D images from natural language descriptions is a key research area in both computer vision and natural language processing. Early methods employed two-stage pipelines, where region proposals were first generated and then matched with linguistic features \cite{zhang2018grounding, hu2017modeling}.
Subsequent advancements introduced tree structures \cite{liu2019learning, hong2019learning} and graph neural networks (GNNs) \cite{yang2019dynamic, wang2019neighbourhood} to better model object relationships. Recently, the trend shifted toward single stage models that directly regressed bounding boxes from fused multi-modal features \cite{sadhu2019zero, yang2019fast}.
Transformer-based methods \cite{du2022visual, deng2021transvg} set new state-of-the-art benchmarks. Moreover, relation-aware grounding techniques, such as expression parsing \cite{chen2022multi, liu2020learning} and graph alignment \cite{wang2019neighbourhood}, further enhanced performance.

\noindent \textbf{3D Visual Grounding.}  
Identifying referred objects in 3D scenes using natural language gained momentum with datasets like ScanRefer \cite{chen2020scanrefer} and SR3D/NR3D \cite{achlioptas2020referit3d}. 
Early methods relied on two-stage pipelines that used pre-trained detectors \cite{qi2017pointnet++} to extract 3D features. 
In contrast, more recent models moved towards transformer-based approaches \cite{zhao20213dvg, he2021transrefer3d}, graph-based techniques \cite{luo20223d}, and unified frameworks tailored for dense 3D tasks \cite{chen2022d, cai20223djcg}. Methods that integrated 2D semantic information \cite{yang2021sat} and efficient single stage grounding models like 3D-SPS \cite{luo20223d} also demonstrated significant improvements. Additionally, research expanded towards outdoor scene grounding \cite{lin2023wildrefer} and cost-effective monocular methods, such as Mono3DVG \cite{zhan2024mono3dvg}, which aimed to reduce reliance on LiDAR.

\noindent \textbf{Multi-Modality Occupancy Perception.}  
Fusing data from multiple sensor modalities has been proven essential for robust occupancy perception, particularly in autonomous driving. Cameras provided rich semantic details, while LiDAR offered precise sparse geometric information. Many methods aligned 2D image features to 3D space and fused them with LiDAR data \cite{wang2023openoccupancy, ming2024occfusion,sze2024real}. Radar integration further improved prediction, as demonstrated by TEOcc \cite{lin2024teocc}. Recent advances incorporated natural language, with models like CLIP \cite{radford2021learning}, POP3D \cite{vobecky2024pop}, and LangOcc \cite{boeder2024langocc} enabling open-vocabulary perception through vision-language alignment. Additionally, OVO \cite{tan2023ovo} aligned voxel predictions with feature maps, though it lacked geometry supervision and was primarily suited for simple scenes.

\begin{table*}[h]
\centering
{
    {\fontsize{6}{0}\selectfont
    \renewcommand{\arraystretch}{1.5}
    \begin{tabular}{@{} l |c c c c c c c c c c c c c c c}
    \toprule
    Objects & \rotatebox[origin=l]{90}{\vphantom{Motorcycle}\crule[carcolor]{0.15cm}{0.15cm} Car} & \rotatebox[origin=l]{90}{\vphantom{Motorcycle}\crule[personcolor]{0.15cm}{0.15cm} Pedestrian} & \rotatebox[origin=l]{90}{\vphantom{Motorcycle}\crule[truckcolor]{0.15cm}{0.15cm} Truck} & \rotatebox[origin=l]{90}{\vphantom{Motorcycle}\crule[bicyclecolor]{0.15cm}{0.15cm} Bicycle} & \rotatebox[origin=l]{90}{\vphantom{Motorcycle}\crule[color1]{0.15cm}{0.15cm} Animal} 
    &\rotatebox[origin=l]{90}{\crule[color2]{0.15cm}{0.15cm} \begin{tabular}{@{}c@{}}Movable\end{tabular}} & 
    \rotatebox[origin=l]{90}{\vphantom{Motorcycle}\crule[trafficsigncolor]{0.15cm}{0.15cm} Tra-Cone.}
    & \rotatebox[origin=l]{90}{\vphantom{Motorcycle}\crule[color3]{0.15cm}{0.15cm} Debris} & \rotatebox[origin=l]{90}{\crule[othervehiclecolor]{0.15cm}{0.15cm} \begin{tabular}{@{}c@{}}Const-Veh.\end{tabular}} & \rotatebox[origin=l]{90}{\vphantom{Motorcycle}\crule[parkingcolor]{0.15cm}{0.15cm} Trailer} & \rotatebox[origin=l]{90}{\vphantom{Motorcycle}\crule[color4]{0.15cm}{0.15cm} Emergency} 
    & \rotatebox[origin=l]{90}{\vphantom{Motorcycle}\crule[bicyclecolor]{0.15cm}{0.15cm} Bike Rack}
    & \rotatebox[origin=l]{90}{\vphantom{Motorcycle}\crule[buildingcolor]{0.15cm}{0.15cm} Barrier} & \rotatebox[origin=l]{90}{\vphantom{Motorcycle}\crule[motorcyclistcolor]{0.15cm}{0.15cm} Motorcycle} & \rotatebox[origin=l]{90}{\vphantom{Motorcycle}\crule[roadcolor]{0.15cm}{0.15cm} Bus} \\ \midrule
    Object Number & 4515 & 2603 & 981 & 246 & 11 & 240 & 254 & 28 & 83 & 124 & 24 & 60 & 166 & 246 & 344 \\
    \midrule
    Average Voxel Number & 180 & 32 & 446 & 54 & 22 & 22 & 36 & 95 & 323 & 772 & 196 & 206 & 39 & 56 & 871 \\
    \bottomrule
    \end{tabular}%
    }
}
\caption{Statistics of referent object count and average number of occupied voxels in Talk2Occ.}
\vspace{-10mm}
\label{table:data_statistics}
\end{table*}
\section{The Talk2Occ Benchmark}
\label{sec:dataset}

\noindent \textbf{Dataset Construction.}  
The Talk2Occ dataset is created by combining Talk2Car \cite{deruyttere2019talk2car} and Occ3D-nuScenes \cite{tian2024occ3d}, both derived from the nuScenes dataset \cite{caesar2020nuscenes}. 
This dataset supports 3D occupancy grounding by linking each Talk2Car sample to a matching nuScenes frame. Using sample tokens as unique identifiers, we retrieve 3D bounding box annotations and occupancy labels from Occ3D-nuScenes, while reusing all 11,959 natural language prompts from Talk2Car. To ensure data quality, we apply two filters: bounding box centers lie within the range of x $\in$ [-40m, 40m], y $\in$ [-40m, 40m], and z $\in$ [-1m, 5.4m], and bounding boxes contain occupied voxels. We also perform a coordinate transformation to align the data with the ego vehicle coordinate system. This process is fully automated without extra manual labeling.

\noindent \textbf{Dataset Statistics.}  
Table \ref{table:data_statistics} shows the distribution of objects and average number of occupied voxels in Talk2Occ, which contains 9,925 objects across 15 categories. Cars make up 45.49\% of instances, followed by pedestrians and trucks. Buses occupy the most voxels, with an average of 871, while smaller objects like pedestrians and traffic cones occupy fewer. The diversity of dataset reflects varying spatial complexities, posing challenges for 3D occupancy grounding. For a robust evaluation, the dataset is split into 8,949 training and 976 validation samples after data filtering.

\noindent \textbf{Metrics.}
We evaluate 3D occupancy grounding using accuracy at different IoU thresholds, specifically Acc@0.25 and Acc@0.5, as established in prior work \cite{chen2020scanrefer, lin2023wildrefer}. For 3D occupancy grounding, we compute the IoU between the ground truth occupancy set $\mathcal{O}_{gt}$ and the predicted occupancy set $\mathcal{O}_{pred}$. The ground truth set $\mathcal{O}_{gt}$ is defined as the occupied voxels $\mathbf{V_{occ}}$ within the referred object’s bounding box $\mathcal{B}$:  
\begin{equation}
\mathcal{O}_{\text{gt}} = \{ v_i \mid v_i \in \mathbf{V}_{\text{occ}} \cap \mathcal{B}, \ell(v_i) \neq \text{free} \}.
\end{equation}

In two-stage methods, $\mathcal{O}_{pred}$ includes the occupied voxels predicted within the predicted bounding box. In single-stage methods, it is directly derived from the model's overall predicted scene occupancy. We calculate the IoU as: 
\begin{equation}
\text{IoU} = \frac{|\mathcal{O}_{gt} \cap \mathcal{O}_{pred}|}{|\mathcal{O}_{gt} \cup \mathcal{O}_{pred}|}.
\end{equation}
The accuracy at a given IoU threshold $\theta$ is defined as:  
\begin{equation}
\text{Acc}@\theta = \frac{1}{N} \sum_{i=1}^{N} \left( \text{IoU}_i > \theta \right),
\end{equation}
where $N$ is the total number of samples. $\left( \text{IoU}_i > \theta \right)$ evaluates to 1 if the IoU for the $i$-th sample exceeds the threshold $\theta$, and 0 otherwise. These metrics provide a comprehensive evaluation of the capability of model to localize and predict 3D occupancy accurately.

\begin{figure*}[!t]
  \centering
  \includegraphics[width=1.0\textwidth]{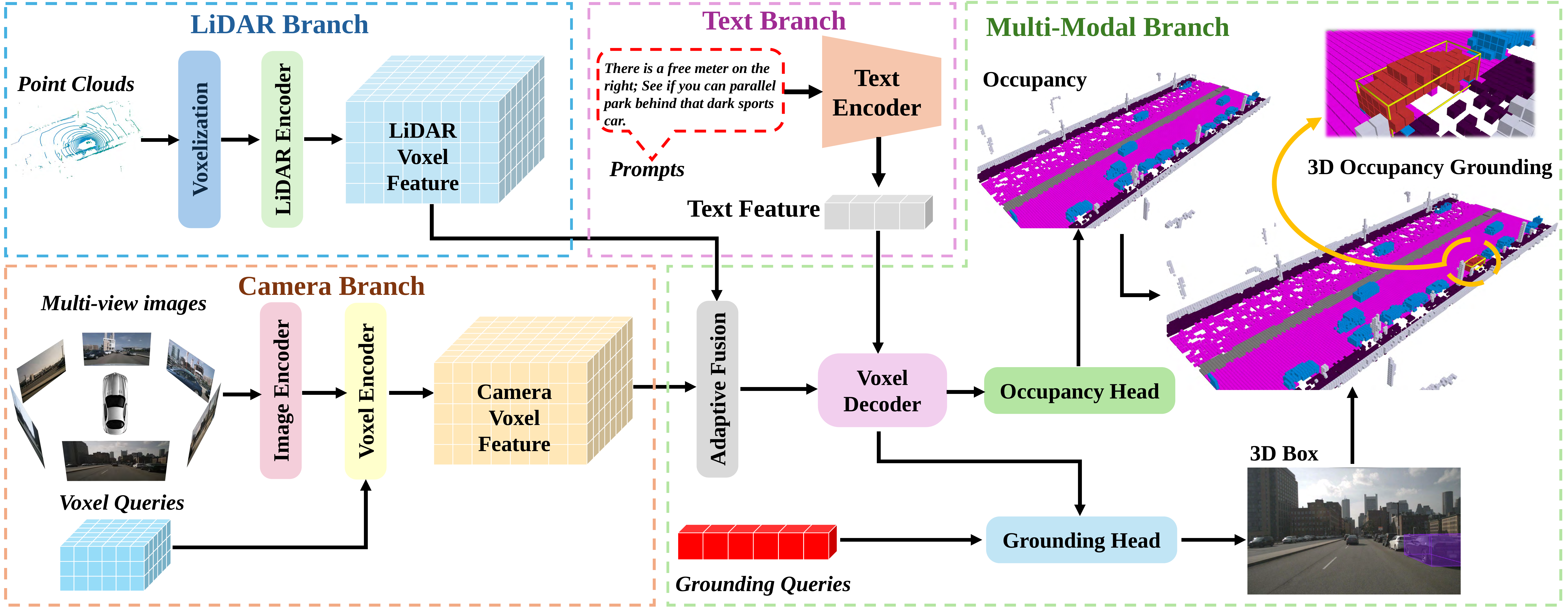}
  \caption{\textbf{Overall architecture of three proposed baselines.} The LiDAR-based baseline (top) employs parameterized voxelization and sparse 3D convolutions to generate LiDAR voxel features, while the camera-based baseline (bottom) utilizes a 3D voxel query mechanism to extract features from multi-view images. The multi-modal baseline combines both modalities through an adaptive fusion module to leverage complementary information.
  }
  \label{fig:baseline}
  \vspace{-6mm}
\end{figure*}
\section{Methodology}
\label{sec:method}
\subsection{Problem Definition}
Multi-modality 3D occupancy grounding is a challenging task in autonomous driving. It involves localizing and inferring the 3D occupancy of a referred object within a scene using a textual prompt $T$, point cloud data $P$, and multi-view images $I$. The prompt $T = \{w_1, \dots, w_L\}$ describes the target object, while $P = \{p_1, \dots, p_N\}$ represents the 3D scene structure. The set of images $I = \{I_1, \dots, I_K\}$ provides multiple perspectives of the scene.
The goal is to predict the occupancy state $\mathcal{O} \in \{0, 1, 2\}^{H \times W \times Z}$ for each voxel. Here, 0 represents free space, 1 indicates a voxel occupied by general objects, and 2 denotes a voxel occupied by the referred object. The dimensions $H$, $W$, and $Z$ correspond to the spatial dimensions of the voxel grid.
Performance is evaluated using the mean Intersection over Union (mIoU) for the voxels occupied by the referred object. This approach allows for a more detailed voxel-level understanding and improves object perception beyond the limitations of traditional bounding box localization.

\subsection{Talk2Occ Baselines}
\label{talk2occ-baseline}
There are currently no methods specifically designed for the 3D occupancy grounding task. Previous language-guided multimodal occupancy perception methods often rely on complex pipelines, first extracting semantic information from images with pre-trained models, then projecting it into 3D space for voxel feature learning \cite{vobecky2024pop, tan2023ovo}. In contrast, we propose three efficient and effective two-stage baseline methods, as illustrated in Fig. \ref{fig:baseline}.
\paragraph{LiDAR-based baseline} 
In the first stage, as shown in the top-left diagram of Fig. \ref{fig:baseline}, we begin by utilizing parameterized voxelization to embed LiDAR points into voxelized features. To enhance computational efficiency, we apply 3D sparse convolutions. This process generates LiDAR voxel features, denoted as $ F^{\mathcal{L}} \in \mathbb{R}^{\frac{H}{S} \times \frac{W}{S} \times \frac{Z}{S} \times D} $, where $ S $ is the stride, leading to a reduction in the spatial dimensions.
Next, a text encoder is used to extract textual features $ F^\mathcal{T} $ from the prompts. These textual features then interact with the voxel features in the voxel decoder. The voxel features are subsequently decoded using 3D convolutions, which produce multi-scale voxel representations.
The occupancy head reduces the feature channels, resulting in the occupancy prediction $ O^\mathcal{L} $. A softplus function \cite{zheng2015improving} is applied to generate semantic probabilities. Additionally, we follow \cite{wang2024panoocc} to learn an occupancy mask $ O^\mathcal{M} \in \{0,1\}^{H \times W \times Z} $ via a binary classification head, which indicates whether positions on $ O^\mathcal{L} $ are occupied. Finally, a grounding query $ Q_{g} $ is used to predict a 3D bounding box $ B $ from the voxel features.

In the second stage, the occupancy predictions are classified. The labels marked as ``free" are denoted as $ O_{free} $, while the remaining labels are classified as $ O_{occ} $. Using the 3D bounding box $ B $, we identify the occupied voxels inside the box, which are marked as $ O_{grounding} $. This leads to the final occupancy grounding prediction, $ O^\mathcal{OG-L} \in \mathbb{R}^{H \times W \times Z} $.

\paragraph{Camera-based baseline}
As depicted at the bottom of Fig. \ref{fig:baseline}, we first employ an image encoder to extract multi-view features $F^{mv}$. Inspired by the querying paradigm in recent works \cite{li2022bevformer, wang2024panoocc}, we introduce a 3D voxel query mechanism($Q_{vq}$). This mechanism extracts features from the multi-view features $F^{mv}$ and prompt embeddings $F^\mathcal{T} \in \mathbb{R}^{L \times d}$. This process outputs camera voxel features $F^\mathcal{C}$, maintaining the same volumetric size as $F^\mathcal{L}$.
Following the LiDAR-based baseline, we apply the voxel decoder, occupancy head, and grounding head to generate the final occupancy grounding results, denoted as $O^\mathcal{OG-C} \in \mathbb{R}^{H \times W \times Z}$.

\makeatletter
\edef\originalParaValue{\the\value{paragraph}} 
\renewcommand{\theparagraph}{\thesubsection.\alph{paragraph}} 
\def\@currentlabel{\theparagraph} 
\paragraph{Multi-modal baseline}
\label{p:multimodal}
The LiDAR voxel features $F^\mathcal{L}$ and camera features $F^\mathcal{C}$ are both effective representations for occupancy prediction. Following \cite{wang2023openoccupancy}, we employ an adaptive model that dynamically integrates these features. The fusion of these features results in combined voxel features $F^\mathcal{F}$. The final occupancy grounding prediction, $O^\mathcal{OG-F}$, is generated using the voxel decoder, occupancy head, and grounding head as described earlier.

To train these baselines, we utilize a combination of loss functions. These include cross-entropy loss $\mathcal{L}_\text{ce}$, focal loss $\mathcal{L}_\text{mask}$ \cite{ross2017focal}, and Lovasz-softmax loss $\mathcal{L}_\text{ls}$ \cite{berman2018lovasz}. Focal loss is applied for classification, and L1 loss is used for bounding box regression, following \cite{li2022bevformer}. The total loss function is then defined as:
\begin{equation}
    \mathcal{L}_\text{baseline} = \mathcal{L}_\text{ce}+ \mathcal{L}_\text{mask} + \mathcal{L}_\text{ls} + \mathcal{L}_\text{cls} + \mathcal{L}_\text{reg}.
\end{equation}
\setcounter{paragraph}{\originalParaValue}
\makeatother

\subsection{GroundingOcc}
\label{groundingocc-method}
\begin{figure*}[h]
  \centering
  \includegraphics[width=0.95\textwidth]{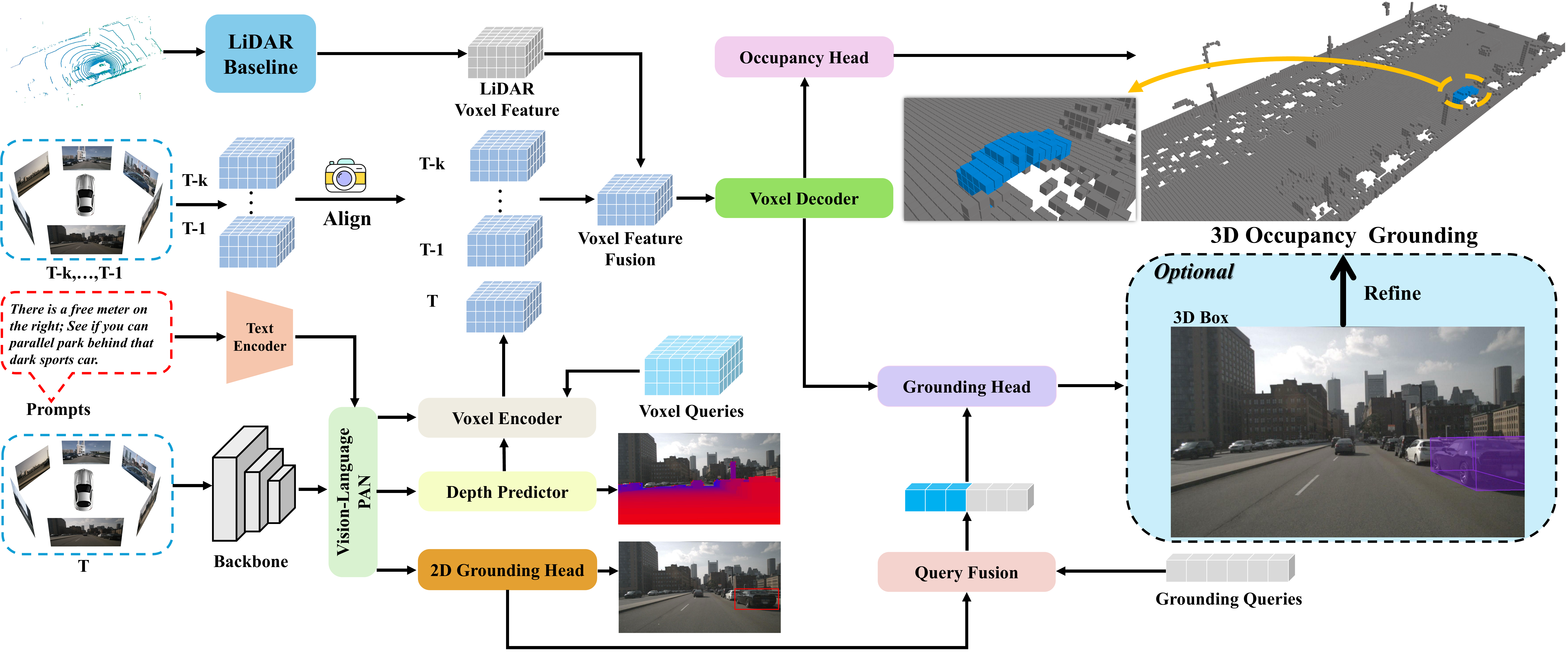}
\caption{\textbf{The overview framework of the multi-modal GroundingOcc.} GroundingOcc replaces semantic supervision with geometric supervision for more precise occupancy grounding prediction, utilizing a voxel encoder and fusion module to generate final occupancy grounding results. The architecture outputs both 3D occupancy grounding predictions and optional 3D bounding boxes that can be refined in post-processing. Here, $T, T-1, \dots, T-k$ denote the current image frame $T$ and its $k$ preceding frames.}
  \label{fig:groundingocc}
  \vspace{-4.5mm}
\end{figure*}

\paragraph{Overview} 
The baseline methods use a two-stage framework, while their complexity in training and deployment presents challenges for practical use. To address these issues, we introduce GroundingOcc, a novel single-stage framework designed for 3D occupancy grounding in autonomous driving. As shown in Fig. \ref{fig:groundingocc}, our approach processes images, point clouds, and text in the same manner as the multi-modal baseline. These multi-modal features are then passed through the Vision-Language PAN module to enhance the interaction between visual and textual representations. To guide feature learning, we incorporate two auxiliary 2D task branches for depth estimation and 2D visual grounding. The fused features are then sent through a voxel encoder and voxel fusion module, generating the final voxel features. The output of GroundingOcc includes the 3D occupancy prediction, along with an optional 3D grounding bounding box that can be refined during post-processing.

\paragraph{Occupancy Encoder}
The occupancy encoder takes in voxel queries $Q_{vq}$, along with the extracted image, text, and point cloud features. It then generates the fused voxel features $F^\mathcal{F}$ in the same manner as above.

\noindent \textbf{Vision-Language Pan.}
Inspired by YOLO-World \cite{cheng2024yolo}, we use the Vision-Language PAN module to create feature pyramids \( \{P_3, P_4, P_5, P_6\} \) from multi-scale image features \( \{C_3, C_4, C_5, C_6\} \). For each scale \( l \in \{3, 4, 5, 6\} \), we aggregate the text embeddings $F^\mathcal{T}$ and image features \( X_l \) using a sigmoid activation function $\sigma(\cdot)$. This is done as follows:
\begin{equation}
X_l' = X_l + X_l \cdot \sigma \left( \max_{j \in \{1, \ldots, L\}} (X_l {F^\mathcal{T}}_j^\top) \right)^\top,
\end{equation}
\noindent \textbf{2D Grounding Head}. At this stage, a transformer decoder is used to process the multi-scale image features \( X_l' \) and the text embedding \( F^\mathcal{T} \). The extracted image features \( X_{2d} \) are defined as below:
\begin{equation}
    X_{2d} =  \text{FFN}(\text{MHCA}(\text{MHSA}(X_l'), F^\mathcal{T})).
\end{equation}
where FFN, MHCA, and MHSA refer to the feed-forward network, multi-head cross-attention, and multi-head self-attention layers, respectively.
Following \cite{carion2020end}, convolutional layers predict the 2D grounding outputs, which include the 2D bounding box coordinates\( (l, r, t, b) \), center \( (x_{2D}, y_{2D}) \), and a centerness measure \( d \).

The integration of 2D spatial locations with multi-scale features and object queries occurs in two stages:

In the first stage, the 2D coordinates \( (H, W) \) in the feature map are normalized as follows:
\begin{equation}
        (x, y)_{\{i,j\}} = \left( \frac{s \cdot i + s/2}{W_{pad}}, \frac{s \cdot j + s/2}{H_{pad}} \right),
\end{equation}
where \( s \) is the stride, and \( (W_{pad}, H_{pad}) \) refer to the padded dimensions of the input image.

In the second stage, the object queries are fused with spatial and feature information using a transformer-based architecture. This fusion process is defined as below:
\begin{equation}
        Q_{\text{fusion}} = \text{Fusion}(Q_{g}, \text{TopK}(F), \text{PE}(p), Q_{pos}).
\end{equation}
where \( Q_{pos} \) and \( Q_{g} \) represent the positional and content components of the grounding queries, respectively. The function \( \text{TopK}(F) \) selects the top-k sampled multi-scale features, while \( \text{PE}(p) \) denotes the position encoding for normalized 3D coordinates.

\begin{figure}[h]
    \centering
    \vspace{2mm}
    \includegraphics[width=0.49\textwidth]{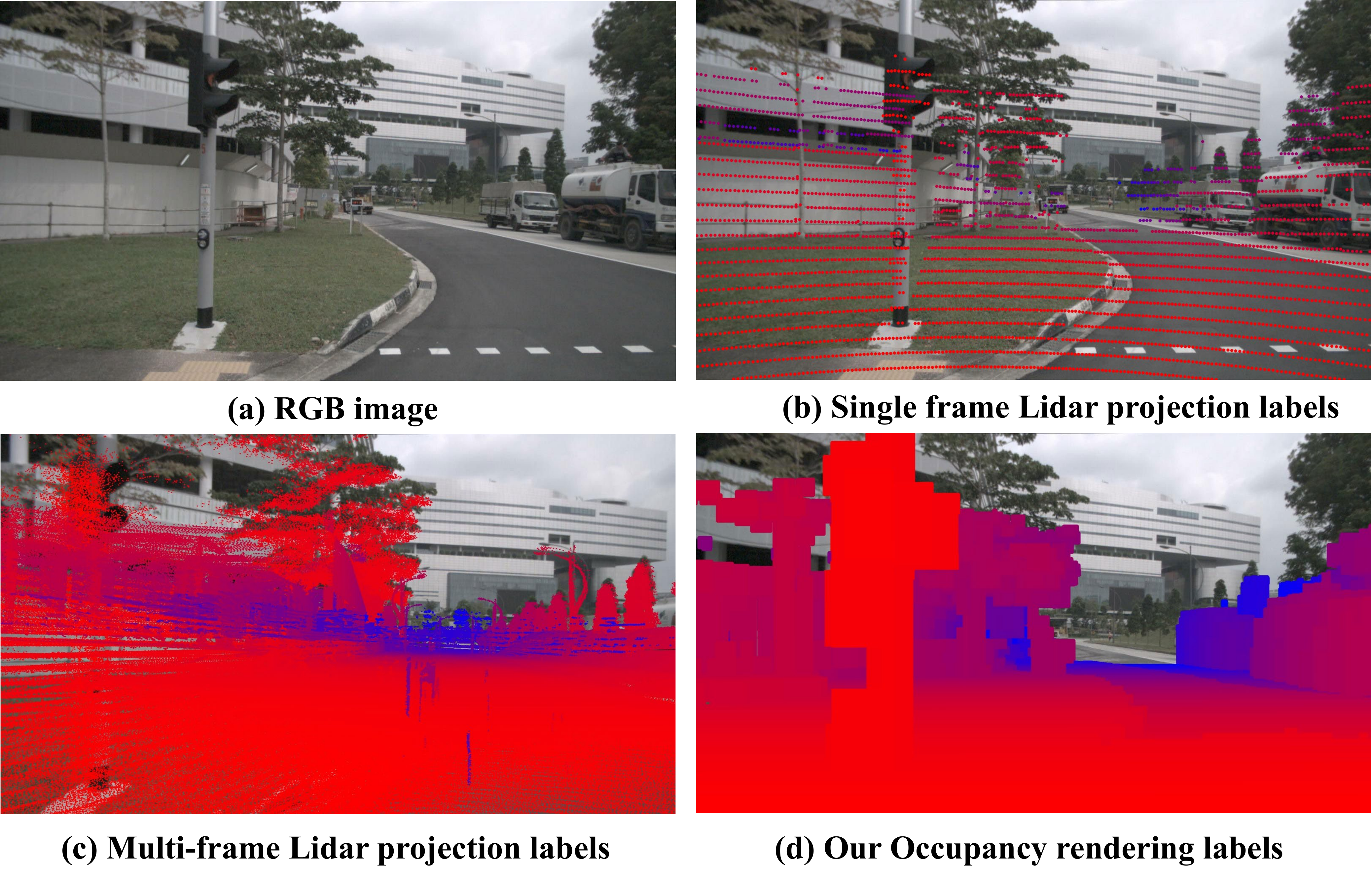}
    \caption{\textbf{Comparison between different depth ground truth generation methods.} Our occupancy-based ray-casting pipeline generates more complete and geometrically accurate depth maps.
    }
    \label{fig:depths}
    \vspace{-4mm}
\end{figure}
\begin{table*}[t!]
\vspace{-2pt}
\centering
\resizebox{\textwidth}{!}{
\begin{tabular}{lccccccc}
\hline
\multirow{2}{*}{{Method}} & \multirow{2}{*}{{Type}} & \multicolumn{2}{c}{{Unique}}  & \multicolumn{2}{c}{{Multiple}} & \multicolumn{2}{c}{{Overall}} \\
 &             & {Acc@0.25(\%)} & {Acc@0.5(\%)} & {Acc@0.25(\%)}  & {Acc@0.5(\%)} & {Acc@0.25(\%)} & {Acc@0.5(\%)} \\ \hline
GT-Rand  & Two-Stage & 4.39 & 3.29 & 4.85 & 3.50 & 4.81 & 3.48\\
Box-Rand & Two-Stage & 5.49 & \underline{5.49} & 5.19 & 4.97 & 5.22 & 5.02 \\ 
\hline
L-baseline& Two-Stage  & 3.29 & 0.00 & 11.86& 1.58 &11.06 & 1.43\\
C-baseline& Two-Stage  & 4.39& 1.09 & 17.17& 1.80 &15.98 & 1.74\\
M-baseline& Two-Stage  & 7.69& 2.19 & 22.48& 2.59 &21.10 & 2.46\\
GroundingOcc& One-Stage & 15.38 & 4.39 & 28.58& 7.68& 27.35& 7.47\\
\rowcolor{blue!7} 
GroundingOcc-Refine& Two-Stage & \textbf{19.78}& \textbf{4.39}&  \textbf{34.01}& \textbf{9.49}& \textbf{32.68}& \textbf{9.01}\\
\hline
\end{tabular}
}
\caption{\textbf{Quantitative results of 3D Occupancy Grounding on Talk2Occ}. The terms "Unique" and "Multiple" are based on whether the object categories referred to in the scene are unique or multiple. The C/L/M-baseline methods are proposed in \ref{talk2occ-baseline}.
}
\label{GroundingOcc_results}
\vspace{-8mm}
\end{table*}

\noindent \textbf{Depth Predictor}. Depth estimation is crucial for 3D object detection and occupancy perception, significantly enhancing representation learning \cite{jiang2024far3d, miao2023occdepth, boeder2024occflownet, zhang2023occnerf}. Traditional methods generate depth ground truth by projecting sparse point clouds onto image coordinates, which often results in inaccuracies due to the sparsity of the LiDAR data.

Therefore, we explore to fuse multi-frame point clouds for denser depth ground truth. However, this introduces spatial misalignment, where distant points may project incorrectly onto nearby regions (see Fig. \ref{fig:depths}(c)). To address this concern, we propose a novel depth ground truth generation method based on occupancy results.
We generate depth maps by converting a 3D occupancy grid, which has the same resolution as in Occ3D\cite{tian2024occ3d}, into camera-view depth maps through ray-casting. For each pixel $(u,v)$, a ray is formed using the inverse camera intrinsic matrix:
\begin{equation}
    [X_c, Y_c, 1]^T = K^{-1}[u, v, 1]^T.
\end{equation}

Next, we march along the ray at fixed intervals $d$ and transform the points to ego coordinates to get $P_{\text{ego}}$:
\begin{equation}
    P_{\text{ego}} = T_{\text{camera\_to\_ego}}[d X_c, d Y_c, d, 1]^T. 
\end{equation}

The ego coordinates are mapped to voxel indices using $
V_{\text{index}} = \left\lfloor \frac{P_{\text{ego}} - P_{\text{min}}}{v_{\text{size}}} \right\rfloor
$, 
where $P_{min}$ is the minimum point cloud bound, and $v_{size}$ is the voxel size. The depth for each pixel is then calculated as the minimum distance along the ray to a non-free voxel:
\begin{equation}
    D(u, v) = \min\{d \mid O(V_{\text{index}}) \neq l_{\text{free}}\}.
\end{equation}

\paragraph{Occupancy Grounding Decoder}
The occupancy grounding decoder takes fused voxel features $F^\mathcal{F}$ and produces occupancy predictions and 3D bounding box estimates through two branches. The grounding head refines 3D occupancy grounding by improving spatial localization.

\noindent \textbf{Occupancy Head}. This module upsamples voxel features $F^\mathcal{F} \in \mathbb{R}^{\frac{H}{S} \times \frac{W}{S} \times \frac{Z}{S} \times D}$ using 3D deconvolution to generate high-resolution occupancy features  $F^\mathcal{O} \in \mathbb{R}^{H \times W \times Z \times D'}$.A multilayer perceptron (MLP) head then predicts occupancy from the derived features $O^\mathcal{OG}$. In addition, a binary voxel mask $O^\mathcal{M}$ is produced to identify the occupied regions. Unlike baseline methods, we use a geometry grounding label for supervision.

\noindent \textbf{Grounding Head}. To predict the 3D bounding box, we first apply average pooling along the height dimension to generate a 2D Bird's Eye View (BEV) embedding $F^\mathcal{BEV}$ from the voxel features $F^\mathcal{O}$:
\begin{equation}
    F^\mathcal{BEV} = \text{Pool}_{\text{avg}}(F^\mathcal{O}).
\end{equation}

We then use a transformer-based decoder, as in \cite{zhu2020deformable}, to predict the 3D bounding box attributes, including the dimensions \( (l, h, w) \), the 3D box center \( (x_{3D}, y_{3D}, z_{3D}) \), and the yaw angle \( \psi \) along the z-axis. The predicted bounding box refines the results of 3D occupancy grounding, similar to baseline methods.

\subsection{Loss Function}
The total loss $\mathcal{L}_\text{ours}$ for training consists of four components: 
\begin{equation}
    \mathcal{L}_\text{ours} = \mathcal{L}_\text{2D}+\mathcal{L}_\text{3D}+\mathcal{L}_{\text{dmap}}+\mathcal{L}_{\text{occ}}.
\end{equation}

The 2D loss \(\mathcal{L}_\text{2D}\) includes box size and center, defined as $\mathcal{L}_\text{2D} = \lambda_1\mathcal{L}_\text{lrtb}+\lambda_2\mathcal{L}_{\text{xy2D}} + \lambda_3\mathcal{L}_{\text{centerness}} + \lambda_4\mathcal{L}_{\text{GIoU}}$. Here, \(\lambda_{1 \sim 4}\) are hyperparameters that are set to (5, 10, 1, 2). The losses \(\mathcal{L}_\text{lrtb}\) and \(\mathcal{L}_\text{xy2D}\) use \(\mathcal{L}_1\) loss, and \(\mathcal{L}_{\text{GIoU}}\) uses GIOU loss. The 3D loss \(\mathcal{L}_\text{3D}\) is the same as baseline methods, for attributes like box size, center, orientation, category, and depth, is defined as $\mathcal{L}_\text{3D} = \lambda_5 \mathcal{L}_{\text{cls}} + \lambda_6 \mathcal{L}_{\text{bbox}}$, where \(\lambda_{5 \sim 6}\) are set to (2, 0.25). For the depth map, we adopt focal loss to supervise \(\mathcal{L}_{\text{dmap}}\). The occupancy loss \(\mathcal{L}_{\text{occ}}\) consists of several terms: $\mathcal{L}_{\text{occ}} = \lambda_7 \mathcal{L}_{\text{ce}} + \lambda_8 \mathcal{L}_{\text{mask}} + \lambda_{9} \mathcal{L}_{\text{Lovasz}} + \lambda_{10} \mathcal{L}_{\text{scal}}^{\text{geo}} + \lambda_{11} \mathcal{L}_{\text{scal}}^{\text{sem}}$, where \(\lambda_{7 \sim 11}\) are set to (10, 2, 1, 1, 1). 
Compared to baseline methods, GroundingOcc adds scene-class affinity losses \(\mathcal{L}_{\text{scal}}^{\text{geo}}\) and \(\mathcal{L}_{\text{scal}}^{\text{sem}}\) from MonoScene\cite{cao2022monoscene}.  
\section{Experiments}
\subsection{Implementation Details}
\begin{figure*}[h]
  \centering
  \includegraphics[width=0.9\textwidth]{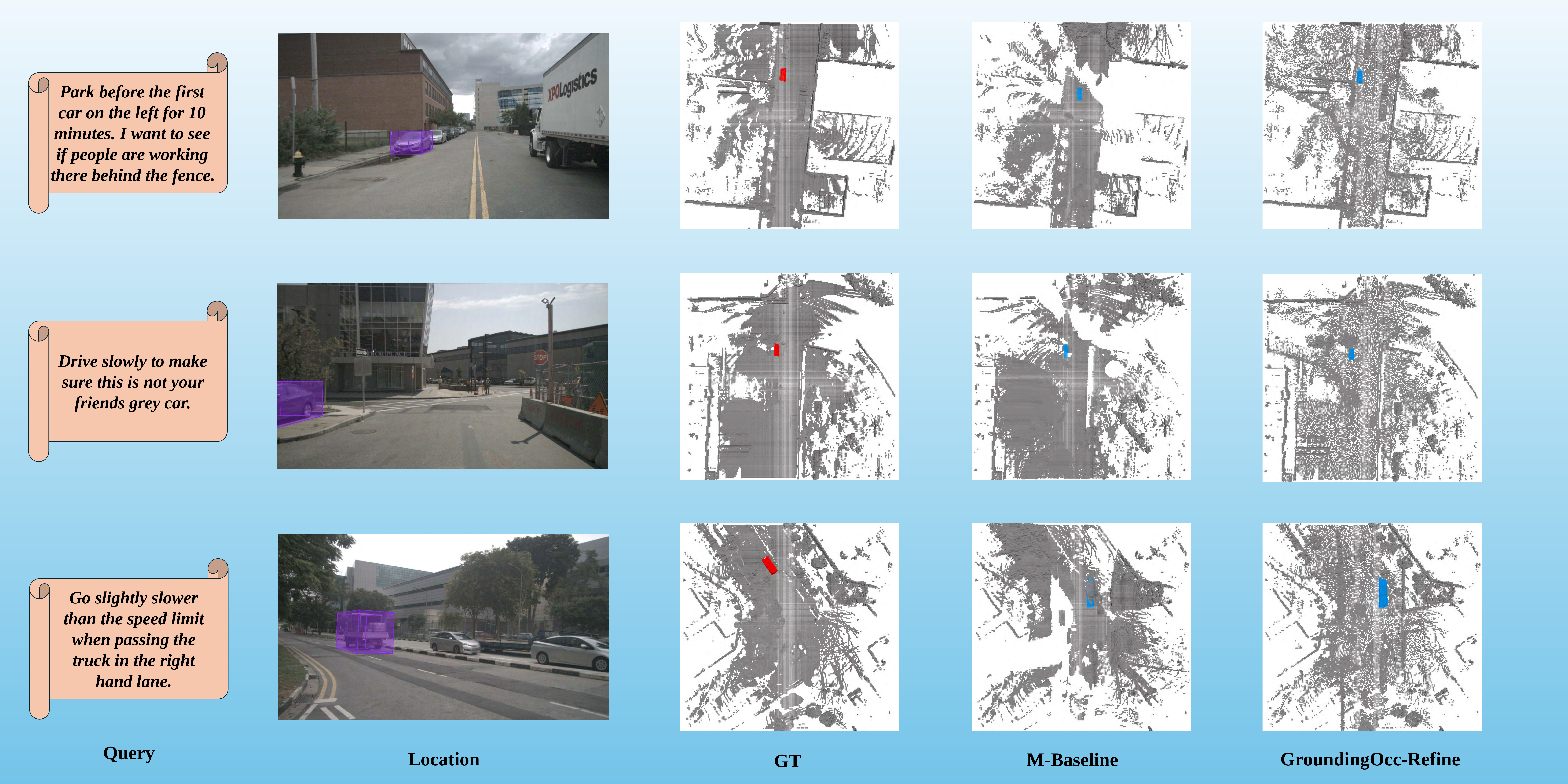}
  \caption{\textbf{Qualitative results from baseline method and our GroundingOcc.} Red, blue, and purple boxes denote the ground truth occupancy grounding, prediction, and ground truth bounding box, respectively. }
  \label{fig:vis}
  \vspace{-5mm}
\end{figure*}
\paragraph{Setting}
Our implementation is based on the nuScenes dataset \cite{caesar2020nuscenes}, like Occ3D-nuScenes \cite{tian2024occ3d}. The point cloud range is \(\left[-40.0\, \text{m}, 40.0\, \text{m}\right]\) for the $x$ and $y$ axes, and \(\left[-1.0\, \text{m}, 5.4\, \text{m}\right]\) for the $z$ axis. We use a voxel grid size of \((0.4\, \text{m}, 0.4\, \text{m}, 0.4\, \text{m})\) for loss supervision and an initial voxel query resolution of \(50 \times 50 \times 16\), with embedding dimension \(D = 256\). After upsampling, the resolution becomes \(200 \times 200 \times 32\) with \(D' = 64\).

For the image backbone, we use ResNet101-DCN \cite{dai2017deformable}, providing multi-scale features at resolutions of $1/8$, $1/16$, $1/32$, and $1/64$. The text encoder is a pre-trained RoBERTa-base model \cite{liu2019roberta}. Depth maps are estimated from the \(P_3\) feature map at $1/8$ resolution. The voxel encoder consists of $3$ layers with $4$ sampling points per voxel query. We fuse data from $4$ consecutive frames ($0.5$s apart) with single-frame LiDAR point clouds to generate fused voxel embeddings. The occupancy head uses two MLP layers with $128$ hidden units and softplus activation, while the grounding head has $3$ layers with grounding queries of dimension $256$.

\paragraph{Training}
We train both the baselines and GroundingOcc model on 4 NVIDIA RTX 4090 GPUs, with a batch size of 1 per GPU. The AdamW optimizer is adopted with an initial learning rate of \(2 \times 10^{-4}\), cosine annealing schedule, and weight decay of $0.01$. Data augmentation techniques such as image cropping, color distortion, and GridMask \cite{chen2020gridmask} are applied. Input images are cropped and resized to \(320 \times 800\) for training due to GPU memory constraints.

\subsection{Main Results}
\paragraph{Quantitative Analysis}
Since no existing methods are specifically designed for the 3D occupancy grounding task, we compare our method with several baselines to demonstrate its effectiveness. (1) \textbf{GT-Rand}: This method randomly selects a ground truth box and generates an occupancy voxel label within the same class range. It serves as a baseline for random performance. (2) \textbf{Box-Rand}: Similar to GT-Rand, it uses ground truth occupancy results. It provides a stronger comparison. (3) \textbf{GroundingOcc}: This is the method we propose in Section \ref{groundingocc-method}. (4) \textbf{GroundingOcc-Refine}: A refined version of GroundingOcc, incorporating a 3D bounding box for more accurate results. The \textbf{C/L/M-baseline} methods are introduced in Section \ref{talk2occ-baseline}.

We report the results on the Talk2Occ dataset in Table \ref{GroundingOcc_results}. Our method, GroundingOcc-Refine, shows significant improvements across all metrics. It achieves 32.68\% Acc@0.25 and 9.01\% Acc@0.5, outperforming strong baseline methods. This confirms the superior performance of our approach in the 3D occupancy grounding task.

\paragraph{Qualitative Analysis}
Figure \ref{fig:vis} presents the 3D occupancy grounding results for M-baseline and GroundingOcc-Refine. The baseline method offers rough estimates of object range and occupancy but struggles to pinpoint precise locations due to limited depth sensitivity. GroundingOcc addresses this by using a depth predictor, improving distance perception. Our method combines a 2D Grounding Head, depth prediction, and multi-frame fusion. This multi-modal approach effectively integrates both appearance and geometric information. However, there are some failure cases. In scenes with ambiguous instructions or multiple similar objects, both the baseline and GroundingOcc methods fail to provide accurate results. 
These cases highlight the challenges posed by complex environments.

\begin{table}[t]
\centering
\renewcommand\tabcolsep{5pt}
\renewcommand{\arraystretch}{1.0}
\scalebox{0.85}{
\begin{tabular}{c|c|c|c|c|c}
\hline
\multirow{2}{*}{\begin{tabular}[c]{@{}c@{}}M-Baseline\end{tabular}} & 
\multirow{2}{*}{\begin{tabular}[c]{@{}c@{}}Multi \\Frames \end{tabular}} & 
\multirow{2}{*}{\begin{tabular}[c]{@{}c@{}}Depth \\Predictor \end{tabular}} &
\multirow{2}{*}{\begin{tabular}[c]{@{}c@{}}2D Grounding \\ Head\end{tabular}} &
\multirow{2}{*}{Acc@0.25} &
\multirow{2}{*}{Acc@0.5} \\ 
 &  & & &  & \\ 
\hline
\checkmark &  &   &   & 21.10  & 2.46  \\
\checkmark &\checkmark    &   &  &21.90  & 3.07  \\
\checkmark & \checkmark    & \checkmark &  &  24.46 &  5.40 \\
\checkmark & \checkmark    & \checkmark & \checkmark &  \textbf{26.94} & \textbf{5.94}  \\
\hline
\end{tabular}
}
\caption{Impact of the proposed module in GroundingOcc framework.}
\label{tab:ablation}
\vspace{-6mm}
\end{table}
\begin{table}[t]
\centering
\renewcommand\tabcolsep{6.5pt}
\renewcommand{\arraystretch}{1.0}
\scalebox{0.9}{
\begin{tabular}{c|c|c|c|c|c}
\hline
\multirow{2}{*}{\begin{tabular}[c]{@{}c@{}}Semantic\\ Supervision\end{tabular}} & 
\multirow{2}{*}{\begin{tabular}[c]{@{}c@{}}Geometric\\ Supervision\end{tabular}} & 
\multirow{2}{*}{$\mathcal{L}_{\text{scal}}^{\text{geo}}$} & 
\multirow{2}{*}{$\mathcal{L}_{\text{scal}}^{\text{sem}}$} & 
\multirow{2}{*}{Acc@0.25} & 
\multirow{2}{*}{Acc@0.5} \\ 
 &  &  &  &  &\\ 
\hline
\checkmark &  &   &   &  26.94 & 5.94  \\
 \checkmark & \checkmark    &   &   & 29.30  & 8.07  \\
 \checkmark & \checkmark    & \checkmark &  &  31.67 &  8.68 \\
\checkmark     & \checkmark    & \checkmark & \checkmark &  \textbf{32.68} & \textbf{9.01}  \\
\hline
\end{tabular}
}
\caption{Ablation study of the occupancy loss function.}
\vspace{-8mm}
\label{tab:ablation_loss}
\end{table}
\subsection{Ablation Study}
\textbf{Impact of Each Module.}
We evaluate the contribution of each component in the GroundingOcc framework using the Talk2Occ validation set. Performance is measured at Acc@0.25 and Acc@0.5. The results are shown in Table \ref{tab:ablation}. Adding multi-frame fusion improves performance slightly. It helps the model capture richer scene information by merging multiple frames. The depth predictor brings a more noticeable improvement, enhancing the model's ability to estimate distances accurately. The 2D grounding head further boosts accuracy. By refining 3D queries with 2D spatial priors, it helps generate better grounding queries.

\textbf{Impact of Occupancy Losses.}
We study the role of different loss components in occupancy learning using the GroundingOcc-Refine model. The results are shown in Table \ref{tab:ablation_loss}. Semantic supervision alone provides a baseline performance. Adding geometric supervision improves accuracy by capturing structural details. The geometric scaling loss further enhances spatial consistency. Finally, incorporating the semantic scaling loss results in the best performance. Notably, a key distinction between GroundingOcc and baseline methods is the use of geometric supervision, which significantly enhances performance by focusing more on geometric quality. 
\section{Conclusion}
\label{sec:conclusion}
In this paper, we introduce Talk2Occ, a new benchmark for multi-modality 3D occupancy grounding in autonomous driving. To tackle this task, we propose GroundingOcc, an end-to-end model that integrates visual, textual, and point cloud features for occupancy grounding. It consists of a multimodal encoder for feature extraction, an occupancy head for voxel-wise predictions, and a grounding head to refine localization. Additionally, a 2D grounding module and a depth estimation module enhance geometric understanding. 
Extensive experiments on Talk2Occ demonstrate that GroundingOcc outperforms the baseline methods, bridging the gap between bounding box-based grounding and voxel-level occupancy grounding.
\section*{Acknowledgments}
This work is supported by National Natural Science Foundation of China under Grant No.62376244.

\bibliographystyle{IEEEtran}
\bibliography{root}

\end{document}